\documentclass[11pt,a4paper]{article}
\usepackage[hyperref]{acl2019}
\usepackage{times}
\usepackage[normalem]{ulem}
\usepackage{latexsym}
\usepackage{url}
\usepackage{adjustbox}
\usepackage{tabularx}
\usepackage{graphics}
\usepackage{multirow}
\usepackage{multicol}
\usepackage{booktabs}
\usepackage{subcaption}
\usepackage{enumitem}
\usepackage[toc,page]{appendix}

\newlist{alphalist}{enumerate}{1}
\setlist[alphalist,1]{label=\textbf{\alph*.}}

\usepackage{pbox}
\usepackage{xfrac}

\usepackage{todonotes}

\aclfinalcopy % Uncomment this line for the final submission
\title{Higher-order Comparisons of Sentence Encoder Representations}
\author{Mostafa Abdou\textsuperscript{$\dagger$}~~~Artur Kulmizev\textsuperscript{$\clubsuit$}~~~Felix Hill\textsuperscript{$\diamondsuit$}~~~Daniel M. Low\textsuperscript{$\spadesuit$}~~~Anders S{\o}gaard\textsuperscript{$\dagger$} \\
  \textsuperscript{$\dagger$}Department of Computer Science, University of Copenhagen~~~{\tt abdou,soegaard@di.ku.dk}\\
  \textsuperscript{$\clubsuit$}Department of Linguistics and Philology, Uppsala University~~~{\tt artur.kulmizev@lingfil.uu.se}\\
  \textsuperscript{$\diamondsuit$}DeepMind~~~{\tt felixhill@google.com}\\
  \textsuperscript{$\spadesuit$}Program in Speech and Hearing Bioscience and Technology, \\Harvard Medical School-MIT~~~{\tt dlow@mit.edu}\\}

\date{}

\begin{document}
\interfootnotelinepenalty=10000

\maketitle

\begin{abstract}

Representational Similarity Analysis (RSA) is a technique developed by neuroscientists for comparing activity patterns of different measurement modalities (e.g., fMRI, electrophysiology, behavior). As a framework, RSA has several advantages over existing approaches to interpretation of language encoders based on probing or diagnostic classification: namely, it does not require large training samples, is not prone to overfitting, and it enables a more transparent comparison between the representational geometries of different models and modalities. We demonstrate the utility of RSA by establishing a previously unknown correspondence between widely-employed pretrained language encoders and human processing difficulty via eye-tracking data, showcasing its potential in the interpretability toolbox for neural models.

\end{abstract}

%\section{Credits}
\section{Introduction}
\label{intro}

Examining the parallels between human and machine learning is a natural way for us to better understand the former and track our progress in the latter. The ``black box'' aspect of neural networks has recently inspired a large body of work related to interpretability, i.e. understanding of representations that such models learn. In NLP, this push has been largely motivated by linguistic questions, such as: \textit{what linguistic properties are captured by neural networks?} and \textit{to what extent do decisions made by neural models reflect established linguistic theories?} Given the relative recency of such questions, much work in the domain so far has been focused on the context of models in isolation (e.g. \textit{what does model} \textbf{X} \textit{learn about linguistic phenomenon} \textbf{Y}?) In order to more broadly understand models' representational tendencies, however, it is vital that such questions be formed not only with other models in mind, but also other representational methods and modalities (e.g. behavioral data, fMRI measurements, etc.). In context of the latter concern, the present-day interpretability toolkit has not yet been able to afford a practical way of reconciling this. 

In this work, we employ Representational Similarity Analysis (RSA) as a simple method of interpreting neural models' representational spaces as they relate to other models and modalities. In particular, we conduct an experiment wherein we investigate the correspondence between human processing difficulty (as reflected by gaze fixation measurements) and the representations induced by popular pretrained language models. In our experiments, we hypothesize that there exists an overlap between the sentences which are difficult for humans to process and those for which per-layer encoder representations are least correlated. 

Our intuition is that such sentences may exhibit factors such as low-frequency vocabulary, lexical ambiguity, and syntactic complexity (e.g. multiple embedded clauses), etc. that are uncommon in both standard language and, relatedly, the corpora employed in training large-scale language models. In the case of a human reader, encountering such a sentence may result in a number of processing delays, e.g. longer aggregate gaze duration. In the case of a sentence encoder, an uncommon sentence may lead to a degradation of representations in the encoder's layers, wherein a lower layer might learn to encode vastly different information than a higher one. Similarly, different models' representations may emphasize different aspects of these more complex sentences and therefore diverge from each other. With this in mind, our hypothesis is that sentences which are difficult for humans to process are likely to have divergent representations within models' internal layers and between different models' layers.

\paragraph{Understanding and analysing language encoders}

In recent years, some prominent efforts towards interpreting neural networks for NLP have included: developing suites that evaluate network representations through performance on downstream tasks \citep{ConneauKSmBB17, wang2018glue, mccann2018natural}; analyzing network predictions on carefully curated datasets \citep{linzen2016assessing, marvin2018targeted, gulordava2018colorless, loula2018rearranging, dasgupta2018evaluating,tenney2018you}; and employing diagnostic classifiers to assess whether certain classes of information are encoded in a model's (intermediate) representations \citep{adi2016fine, chrupala2017representations, hupkes2017visualisation, belinkov2017evaluating}. 

% Much of this work relates to probing pretrained language encoders for linguistic knowledge.

%This latter method has become of particular interest in recent times, shedding valuable insight into how neural networks process filler-gap dependencies, hierarchical structure, among many other phenomena. 
\begin{figure*}[t]
    \centering
    \includegraphics[scale=0.23]{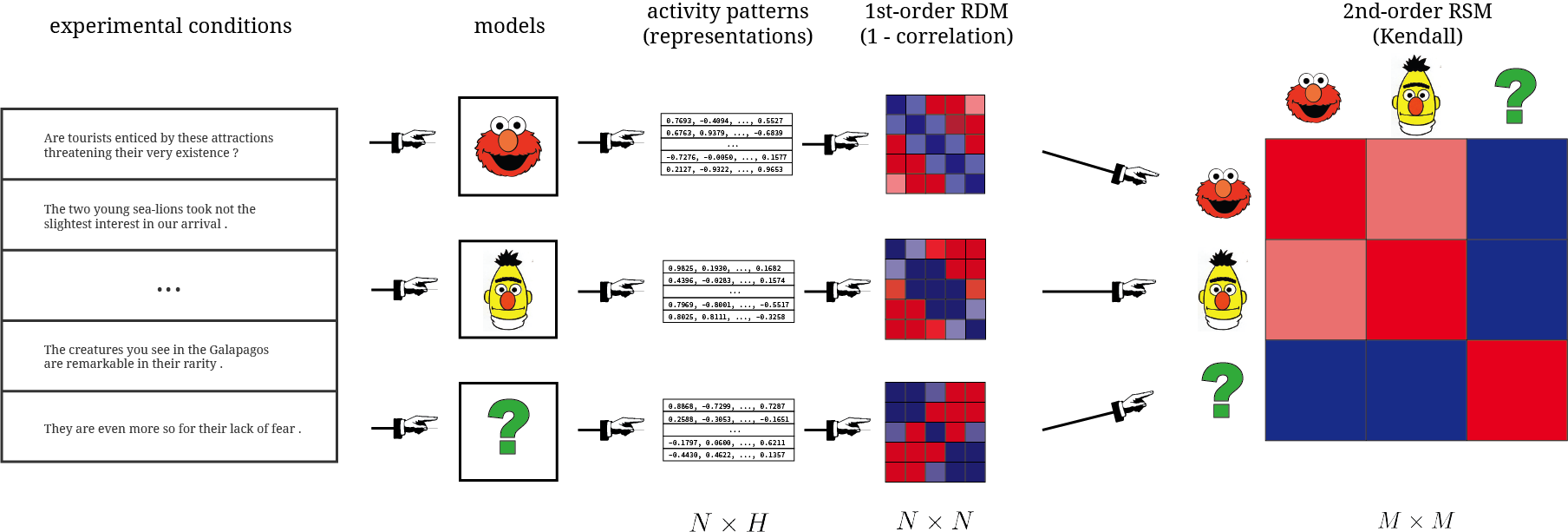}
    %\vspace{-4mm}
    \caption{An example of first- and second-order analyses, where $N = \#$ of experimental conditions, $M = \#$ of models, and $H = \#$ of activity patterns observed for a given model (i.e. dimensionality). The right-most side of the figure depicts a representational similairty matrix (RSM) of correlations between RDMs.}
    \label{fig:rsa_final}
\end{figure*}

While these approaches provide valuable insights into how neural networks process a large variety of phenomena, they rely on decoding accuracy as a probe for encoded linguistic information. If properly biased, this means that they can detect whether information is encoded in a representation or not. However, they do not allow for a direct comparison of representational structure between models. Consider a toy dataset of five sentences of interest and three encodings derived from quite different processing models; a hidden state of a trained neural language model, a \emph{tf-idf} weighted bag-of-words representation, and measurements of fixation duration from an eye-tracking device. Probing methods do not allow us to quantify or visualise, for each of these encoding strategies, how the encoder's responses to the five sentences relate to each other. Moreover, probing methods would not directly reveal whether the fixations from the eye-tracking device aligned more closely with the tf-idf representation or the states of the neural language model. In short, while probing classifier methods can establish if phenomena are separable based on the provided representations, they do not tell us about the overall geometry of the representational spaces. RSA, on the other hand, provides a basis for higher-order comparisons between spaces of representations, and a way to visualise and quantify the extent to which they are isomorphic. 

Indeed, RSA has seen a modest introduction within interpretable NLP in recent years. For example, \citet{chrupala2017representations} employed RSA as a means of correlating encoder representations of speech, text, and images in a post-hoc analysis of a multi-task neural pipeline. Similarly, \citet{bouchacourt-baroni-2018-agents} used the framework to measure the similarity between input image embeddings and the representations of the same image by an agent in an language game setting. More recently, \citet{chrupala2019correlating} correlated activation patterns of sentence encoders with symbolic representations, such as syntax trees. Lastly, similar to our work here,  \citet{abnar-etal-2019-blackbox} proposed an extension to RSA that enables the comparison of a single model in the face of isolated, changing parameters, and employed this metric along with RSA to correlate NLP models' and human brains' respective representations of language. We hope to position our work among this brief survey and further demonstrate the flexibility of RSA across several levels of abstraction.

\section{Representational Similarity Analysis}%\todo{DL: Shouldnt title be "Method: Rep Sim Anal?"}
\label{RSA}
RSA was proposed by \citet{kriegeskorte2008representational} as a method of relating the different representational modalities employed in neuroscientific studies. Due to the lack of correspondence between the activity patterns of disparate measurement modalities (e.g. brain activity via fMRI, behavioural responses), RSA aims to abstract away from the activity patterns themselves and instead compute representational dissimilarity matrices (RDMs), which characterize the information carried by a given representation method through dissimilarity structure. 

Given a set of representational methods (e.g., pretrained encoders) %from different spaces or modalities 
$M$ and a set of experimental conditions (sentences) $N$, we can construct RDMs for each  method in $M$. Each cell in an RDM corresponds to the dissimilarity between the activity patterns associated with pairs of experimental conditions $n_i, n_j \in N$, say, a pair of sentences. When $n_i = n_j$, the dissimilarity between an experimental condition and itself is intuitively $0$, thus making the $N \times N$ RDM symmetric along a diagonal of zeros \citep{kriegeskorte2008representational}. %These RDMs serve as the signatures of a representational method which abstract from their spatial layout; they are indexed on the $x$ and $y$ axes by experimental condition. 

The RDMs of the different representational methods in $M$ can then be directly compared in a Representational \textit{Similarity} Matrix (RSM). This comparison of RDMs is known as second-order analysis, which is broadly based on the idea of a \textit{second-order isomorphism} \citep{shepard1970second}. In such an analysis, the principal point of comparison is the match between the dissimilarity structure of the different representational methods. Intuitively, this can be expressed through the notion of \textit{distance between distances}, and is thus related to Earth Mover's Distance \cite{Rubner:ea:00}.\footnote{More precisely, our measure of dissimilarity between experimental conditions is analogous to \textit{ground distance} and dissimilarity between RDMs to \textit{earth mover's distance.}} Figure \ref{fig:rsa_final} shows an illustration of the first and second order analyses for pretrained language encoders.

Note that RSA is meaningfully different from, and complementary to, methods that employ saturating functions of representation distances (e.g. decoding accuracy, mutual information), which suffer from \textbf{(a)} a ceiling effect: being able to distinguish experimental phenomenon $A$ from $B$ with with an accuracy of 100\% and experimental phenomenon $C$ from $D$ with an accuracy of 100\% does not mean that the distance between $A$ and $B$ is the same as that between $C$ and $D$; and \textbf{(b)} discretization \cite{nili2014toolbox}.

We follow  \citet{kriegeskorte2008representational} in using the correlation distance of experimental condition pairs $n_i, n_j \in N$ as a dissimilarity measure, where $\bar{n_i}$ is the mean of $n_i$'s elements, $\cdot$ is the dot product, and $\|$ is the $l_2$ norm: $corr(x) = 1 - \frac{(n_i - \bar{n_i}) \cdot (n_j - \bar{n_j}) }{{\|(n_i - \bar{n_i}\|}_2 {\|(n_j - \bar{n_j}\|}_2}$.
Compared to other measures, correlation distance is preferable as it normalizes both the mean and variance of activity patterns over experimental conditions. Other popular measures include the Euclidean distance and the Malahanobis distance \citep{kriegeskorte2006information}.

\section{Fixation Duration and Encoder Disagreement}

% \paragraph{Gaze fixation and processing difficulty}
Gaze fixation patterns have been shown to strongly reflect the online cognitive processing demands of human readers \cite{raney2014using, ashby2005eye} and to be dependent upon a number of linguistic factors \cite{van2007eye}. Specifically, it has been demonstrated that word frequency, syntactic complexity, and lexical ambiguity play a strong part in determining which sentences are difficult for humans to process \cite{rayner1986lexical, duffy1988lexical, levy2008expectation}. 

Using the RSA framework, we aim to explore how gaze fixation patterns and the linguistic factors associated with sentence processing difficulty relate to the representational spaces of popular language encoders. Namely, we hypothesize that, for a given sentence, disagreement between hidden layers corresponds to processing difficulty. Because layer disagreement for a sentence measures the extent to which two layers (e.g. within BERT) disagree with each other about the pairwise similarity of the sentence (with other sentences in the corpus), a sentence with high layer disagreement will have unstable similarity relationships to other sentences in the corpus. This indicates that it has a degraded encoder representation. Going further, we also hypothesize that models' representations of said sentences may be confounded, in part, by factors that are known to influence humans. 

\paragraph{Eye-tracking data} For our experiments, we make use of the Dundee eye-tracking corpus \cite{kennedy2003dundee}, the English part of which consists of eye-movement data recorded as 10 native participants read 2,368 sentences from 20 newspaper articles. %published in the \textit{The Independent}. 
We consider the following fixation features: \textsc{Total fixation duration} and \textsc{First Pass duration}. For each of the features, we first take the average of the measurements recorded for all 10 participants per word, then obtain sentence-level annotations by summing the measurements of all words in a sentence and dividing by its length. The result of this is two vectors $V_{totfix}$ and $V_{firstpass}$ of length $2,368$, where each cell in the vector corresponds to a sentence's average total fixation and average first pass duration, respectively.

\paragraph{Syntactic complexity, word frequency, and lexical ambiguity}
We also consider the three following linguistic features which affect processing difficulty. For each of the following the result is also a vector of length $2,368$ where each cell corresponds to a sentence:

\begin{alphalist}
\setlength{\itemsep}{0.5mm}
\setlength{\parskip}{.1mm}
\setlength{\parsep}{0mm}
\item the average word log frequency per sentence extracted from the British National Corpus \cite{leech1992100}, $V_{logFreq.}$. 
\item the average number of senses per word per sentence extracted from WordNet \cite{miller1995wordnet}, $V_{word Sense}$.

\item Yngve scores, a standard measure of syntactic complexity based on cognitive load \cite{yngve1960model} , $V_{Yngve}$. 
\end{alphalist}

\paragraph{Pretrained encoders} We conduct our analysis on pretrained BERT-large \cite{devlin2018bert} and ELMo \cite{peters2018deep}, two widely employed contextual sentence encoders.  
To obtain a representation of a sentence from a given layer $L$, we perform mean-pooling over the time-steps which correspond to the words of a sentence, obtaining a vector representation of the sentence. Mean-pooling is a common approach for obtaining vector representations of sentences for downstream tasks \cite{peters2018deep, conneau2017supervised}. We refer to ELMo's lowest layer as \textsc{E1}, BERT's 11th layer as \textsc{B11}, etc.

\begin{figure*}[ht!]
    \centering
    \begin{subfigure}[t]{0.49\textwidth}
        \centering
        \includegraphics[scale=0.28]{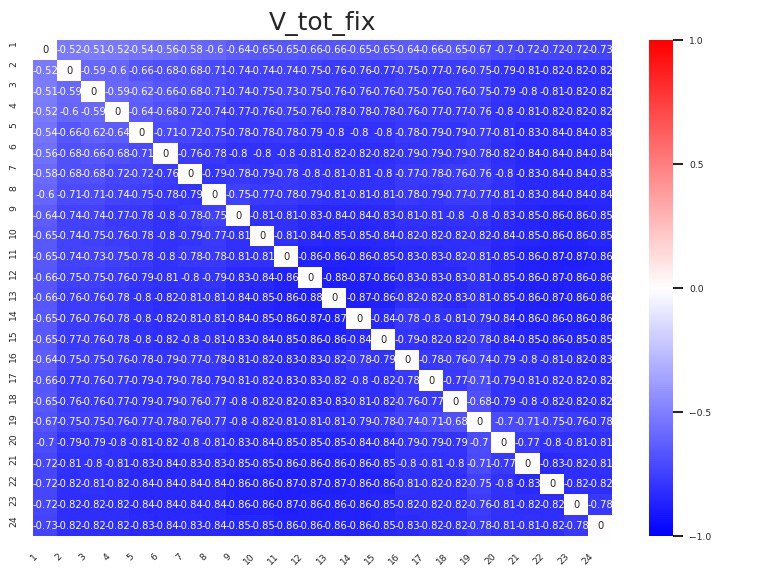}
        \vspace{-4mm}
        \label{fig:mling1st}
    \end{subfigure}
    \begin{subfigure}[t]{0.49\textwidth}
        \centering
        \includegraphics[scale=0.28]{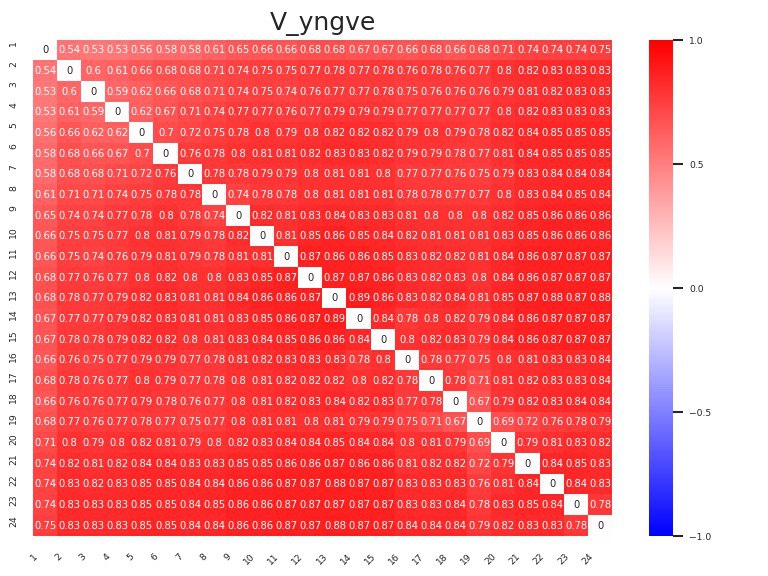}
        \vspace{-4mm}
        \label{fig:mling2nd}
    \end{subfigure}
    \vspace{-4mm}
    \caption{RSMs showing (Spearman's $\rho$) correlation between  disagreement among layers $i$ and $j$ ($V_{Corr_{L_i-L_j}}$) and $V_{totfix}$ (left) and $V_{Yngve}$ (Right). BERT layers are denoted with numbers from 1 (topmost) to 24 (lowest). }
    \label{fig:heatmaps}
\end{figure*}

\paragraph{RDMs} We construct an RDM (see \S\ref{RSA}) for each contextual encoder's layers. Each RDM is a $2,368$ $\times$ $2,368$ matrix which represents the dissimilarity structure of the layer, (i.e., each row vector in the matrix contains the dissimilarity of a given sentence to every other sentence). We then compute the correlations between the two different RDMs. For our evaluation of how well the representational geometry of a layer correlates to another, we employ Kendall's $\tau_{A}$ as suggested in \newcite{nili2014toolbox}, computing the pairwise correlation for each two corresponding rows in two RDMs. This second-order analysis gives us a pairwise relational similarity vector $V_{Corr_{L_i-L_j}}$ of length $2,368$, which has the correlations between two layers \texttt{$L_i$} and \texttt{$L_j$}'s RDMs for each of the sentences. 

\paragraph{Third-order analysis} The final part of our analysis involves computing correlations (Spearman's $\rho$) of  $\{V_{Corr_{L_i-L_j}},V_{logFreq},V_{Yngve},  V_{word Sense}\}$ with each of $V_{totfix}$ and $V_{firstpass}$. The results from this are shown in Table \ref{tab:results1}. The top section of the table shows correlations when \texttt{$L_i$} and \texttt{$L_j$} are the three final adjacent layers in BERT and ELMo. The middle section shows the results for top three BERT layer pairs \texttt{$L_i$} and \texttt{$L_j$} which maximize the correlation scores. The final section shows correlation with the linguistic features. Finally, Figure \ref{fig:heatmaps} shows Spearman's $\rho$ correlations between $V_{Corr_{L_i-L_j}}$ and each of $V_{totfix}$, and $V_{Yngve}$ for all combinations of the 24 BERT layers.

\section{Discussion}
\label{discussion}

Our results show highly significant negative correlations between $V_{Corr_{L_i-L_j}}$ and sentence gaze fixation times. These findings confirm the hypothesis that the sentences that are most challenging for humans to process, are the sentences \textbf{(a)} the layers of BERT disagree most on among themselves; and \textbf{(b)} that ELMo and BERT disagree most on, indicating that there may be common factors which affect human processing difficulty and result in disagreement between layers. By Layer disagreement we refer to the expression $1 - V_{Corr_{L_i-L_j}}$. It is important to note that these encoders are trained with a language modelling objective, unlike models where reading behaviour is explicitly modelled \cite{hahn2016modeling} or predicted \cite{matthies2013blinkers}. Indeed, the similarities here emerge naturally as a function of the task being performed. This can be seen as analogous to the case of similarities observed between neural networks trained to perform object recognition and spatio-temporal cortical dynamics \cite{cichy2016comparison}. 

\begin{table}[ht!]
\centering
\begin{adjustbox}{width=\linewidth}
\begin{tabular}{rcc}
\toprule
\textbf{Layer Disagreement}
& \textbf{Total Fixation}  & \textbf{First Pass Duration} \\
\midrule

E1-B22 & -0.46 & -0.46 \\ 
E2-B23 &   -0.66  & -0.67 \\
E3-B24 &  -0.22  & -0.23  \\
\midrule
B11-B12 &  -0.88   & -0.87  \\
B12-B13 &  -0.87  & -0.85  \\
B10-B21 &  -0.87  &  -0.86 \\
\toprule
\textbf{Linguistic Features}  \\
\midrule

Log Freq.  & -0.20 & -0.19 \\
Avg. Senses per Word & -0.007* & -0.004*  \\
%ngve Score & -0.65 & -0.66 \\
Yngve Score & 0.66 & 0.66 \\
\bottomrule 
\end{tabular}
\end{adjustbox}
\caption{Spearman's $\rho$ between $V_{Corr_{L_i-L_j}}$, $V_{log Freq.}$, $V_{word Sense}$, $V_{Yngve}$ and each of $V_{totfix}$ and $V_{firstpass}$. All correlations significant with $p < 0.0001$ after Bonferroni correction unless marked with *.}
\label{tab:results1}
\end{table}

\paragraph{Syntactic complexity}
Figure \ref{fig:heatmaps} shows that, for all combinations of BERT layers, total fixation time and Yngve scores have strong negative and positive correlations (respectively) with layer disagreement. Furthermore, we observe that disagreement between middle layers seems to show the strongest correlation with Yngve scores. To confirm this, we split the correlations into four groups: ``low'' ($i, j \in [1,8]$), ``middle'' ($i, j \in [9,16]$), ``high'' ($i, j \in [17,24]$), and ``out'' ($|i-j| > 7$), with the latter representing out-of-group correlations (e.g. $Corr_{L_1-L_{24}}$). To account for correlations between disagreeing adjacent layers (e.g. $|i-j| = 1$) and Yngve scores being higher (as a possible confounding factor), we also distinguish layers as either ``adjacent'' or ``non-adjacent''. Considering these two factors as three- and two-leveled independent variables respectively, we conduct a two-way analysis of variance. The analysis reveals that the effect of group is significant at $F(3,275) = 78.47, p < 0.0001$, with ``low'' ($\mu$ = 0.65, $\sigma$ = 0.08), ``middle'' ($\mu$ = 0.84, $\sigma$ = 0.03), ``high'' ($\mu$ = 0.80, $\sigma$ = 0.05), and ``out'' ($\mu$ = 0.80, $\sigma$ = 0.05). Neither the effect of adjacency nor its interaction with group proved to be significant.

This can be seen as (modest) support for the findings of previous work \cite{blevins2018deep, tenney2019bert}: namely, that the intermediate layers of neural language models encode the most syntax, and are therefore possibly more sensitive towards syntactic complexity. A very similar pattern is observed for total fixation time. When considered together with the correlation between $V_{Yngve}$ and fixation times, this indicates a tripartite affinity between layer disagreement, syntactic complexity, and fixation. 

\paragraph{Lexical Ambiguity and Word Frequency} Finally, we observe that $V_{log Freq.}$ has a moderate correlation with both fixation time and layer disagreement and that $V_{word Sense}$ is nearly uncorrelated to both. Detailed plots of the latter can be found in Appendix \ref{sec:supplementalA}. 

\section{Conclusion}
We presented a framework for analyzing neural network representations (RSA) that allowed us to relate human sentence processing data with language encoder representations. In experiments conducted on two widely used encoders, our findings show that sentences which are difficult for humans to process have more divergent representations both intra-encoder and between different encoders. Furthermore, we lend modest support to the intuition that a model's middle layers encode comparatively more syntax. Our framework offers insight that is complimentary to decoding or probing approaches, and is particularly useful to compare representations from across modalities.   
\label{conclusion}

\section*{Acknowledgements} We would like to thank Vinit Ravishankar, Matt Lamm, and the anonymous reviewers for their helpful comments. Mostafa Abdou and Anders S{\o}gaard are supported by a Google Focused Research Award and a Facebook Research Award. 

\bibliographystyle{apalike}
\bibliography{acl2019.bib}

\appendix

\section{Correlation Heatmaps}
\label{sec:supplementalA}
\begin{figure*}
    \centering
    \begin{subfigure}[t]{0.6\textwidth}
        \centering
        \includegraphics[scale=0.38]{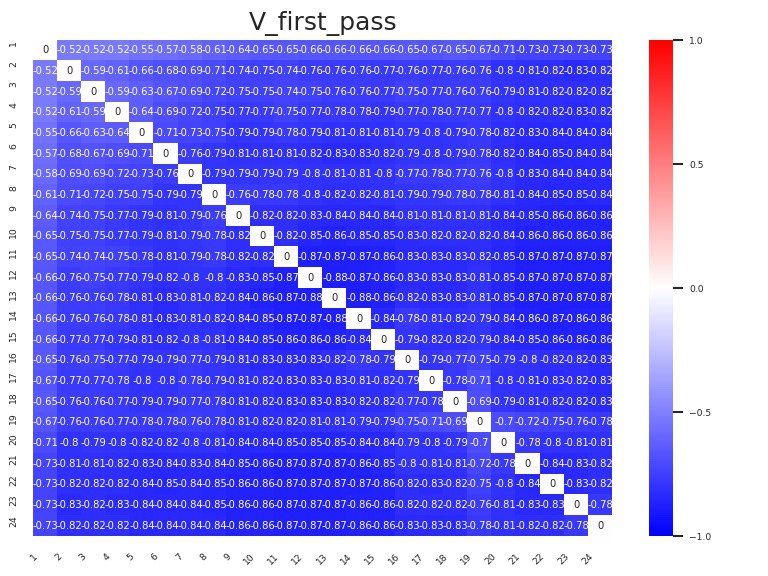}
        \vspace{-4mm}
        \label{fig:first_fix}
    \end{subfigure}
    \begin{subfigure}[t]{0.6\textwidth}
        \centering
        \includegraphics[scale=0.38]{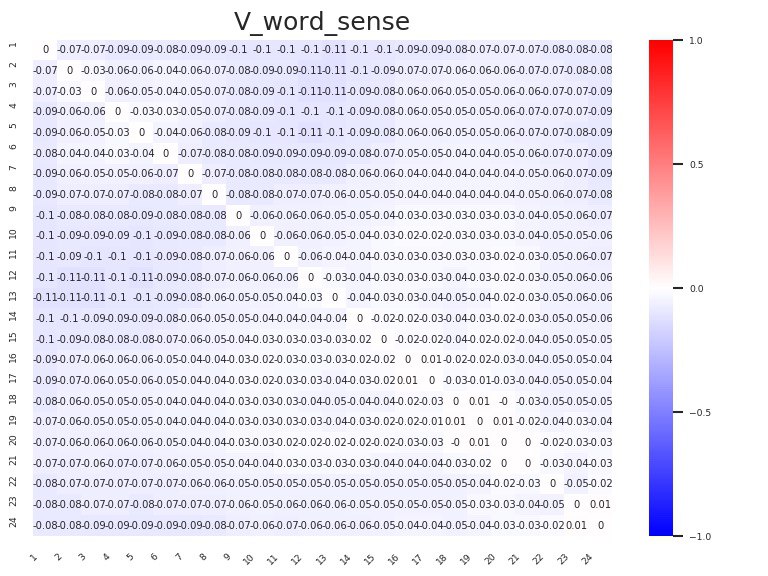}
        \vspace{-4mm}
        \label{fig:word_sense}
    \end{subfigure}
    \begin{subfigure}[t]{0.6\textwidth}
        \centering
        \includegraphics[scale=0.38]{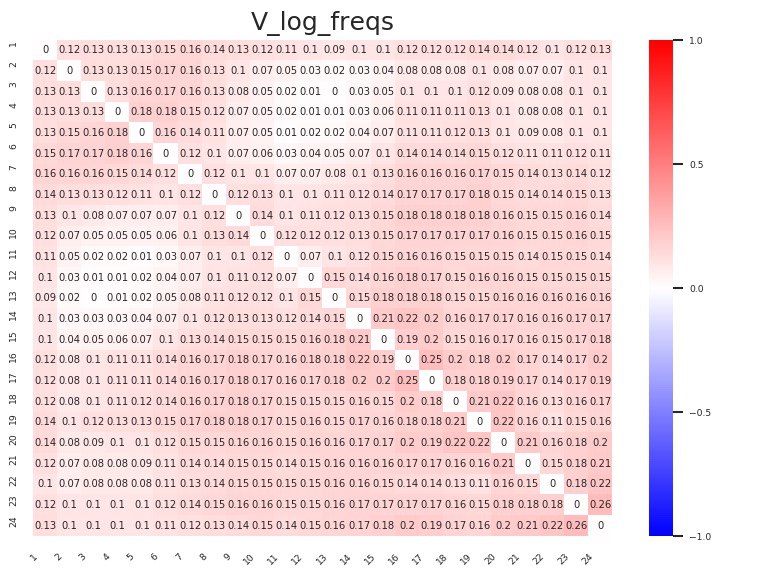}
        \vspace{-4mm}
        \label{fig:word_freq}
    \end{subfigure}
    \vspace{-4mm}
    \caption{RSM heatmaps showing (Spearman's $\rho$) correlation between  disagreement among layers $i$ and $j$ ($V_{Corr_{L_i-L_j}}$) and \textbf{(a)} $V_{firstpass}$ (top), \textbf{(b)} $V_{wordSense}$ (middle) and, \textbf{(c)} $V_{logFreq}$ (bottom).}
    \label{fig:heatmaps_appendix}
\end{figure*}

\end{document}